\documentclass[letterpaper]{article} 
\usepackage{aaai2026}  
\usepackage{times}  
\usepackage{helvet}  
\usepackage{courier}  
\usepackage[hyphens]{url}  
\usepackage{graphicx} 
\usepackage{natbib}  
\usepackage{caption} 
\usepackage{algorithm}
\usepackage{algorithmic}

\usepackage{multirow}
\usepackage{amsmath}
\usepackage{booktabs}

\usepackage{times}
\usepackage{helvet}
\usepackage{courier}
\usepackage{xcolor}
\usepackage{newfloat}
\usepackage{listings}
\DeclareCaptionStyle{ruled}{labelfont=normalfont,labelsep=colon,strut=off} 
\floatstyle{ruled}
\newfloat{listing}{tb}{lst}{}
\floatname{listing}{Listing}
\title{4KDehazeFlow: Ultra-High-Definition Image Dehazing via Flow Matching}
\author{
	Xingchi Chen\textsuperscript{\rm 1}, 
    Pu Wang\textsuperscript{\rm 2}, 
	Xuerui Li\textsuperscript{\rm 3},
	Chaopeng Li\textsuperscript{\rm 4},\\ 
	Juxiang Zhou\textsuperscript{\rm 5}, 
	Jianhou Gan\textsuperscript{\rm 5}, 
     Dianjie Lu\textsuperscript{\rm 6},
    Guijuan Zhang\textsuperscript{\rm 6},
	Wenqi Ren\textsuperscript{\rm 1},
    Zhuoran Zheng\textsuperscript{\rm 8}\thanks{Corresponding Author.}
}
\affiliations{
\textsuperscript{\rm 1}Shenzhen Campus of Sun Yat-sen University\\ 
\textsuperscript{\rm 2}Shandong University\\ 
\textsuperscript{\rm 3}The State University of New York at Buffalo\\ 
\textsuperscript{\rm 4}Jimei University\\
\textsuperscript{\rm 5}the Faculty of Education at Yunnan Normal University\\
\textsuperscript{\rm 7}Shandong Normal University\\
\textsuperscript{\rm 8}Qilu University of Technology\\
}

\usepackage{bibentry}

\begin{document}

\maketitle

\begin{abstract}
	Ultra-High-Definition (UHD) image dehazing faces challenges such as limited scene adaptability in prior-based methods and high computational complexity with color distortion in deep learning approaches. 
	To address these issues, we propose \textbf{4KDehazeFlow}, a novel method based on Flow Matching and the Haze-Aware vector field. 
	This method models the dehazing process as a progressive optimization of continuous vector field flow, providing efficient data-driven adaptive nonlinear color transformation for high-quality dehazing. 
	Specifically, our method has the following advantages: 
	1) 4KDehazeFlow is a general method compatible with various deep learning networks, without relying on any specific network architecture. 
	%
	2) We propose a learnable 3D lookup table (LUT) that encodes haze transformation parameters into a compact 3D mapping matrix, enabling efficient inference through precomputed mappings. 
	3) We utilize a fourth-order Runge-Kutta (RK4) ordinary differential equation (ODE) solver to stably solve the dehazing flow field through an accurate step-by-step iterative method, effectively suppressing artifacts. 
	%
	Extensive experiments show that 4KDehazeFlow exceeds seven state-of-the-art methods. It delivers a 2dB PSNR increase and better performance in dense haze and color fidelity.
\end{abstract}


\section{Introduction}
\label{sec:intro}
With the widespread adoption of 4K imaging devices, existing dehazing methods face challenges in resolution adaptability. 
%
Traditional physical model-based methods produce color distortion and artifacts under complex atmospheric conditions, failing to meet UHD texture requirements. 
Most deep learning-based dehazing methods cannot infer a 4K resolution image on resource-constrained devices.
%
More critically, existing methods are often forced to reduce the model capacity for efficiency, resulting in incomplete haze estimation and lower accuracy in downstream tasks like object detection and semantic segmentation. 
Therefore, developing a UHD dehazing method that balances perceptual quality and computational efficiency remains an urgent challenge in computer vision.

\begin{figure}[t]
	\centering
	\includegraphics[width=\columnwidth]{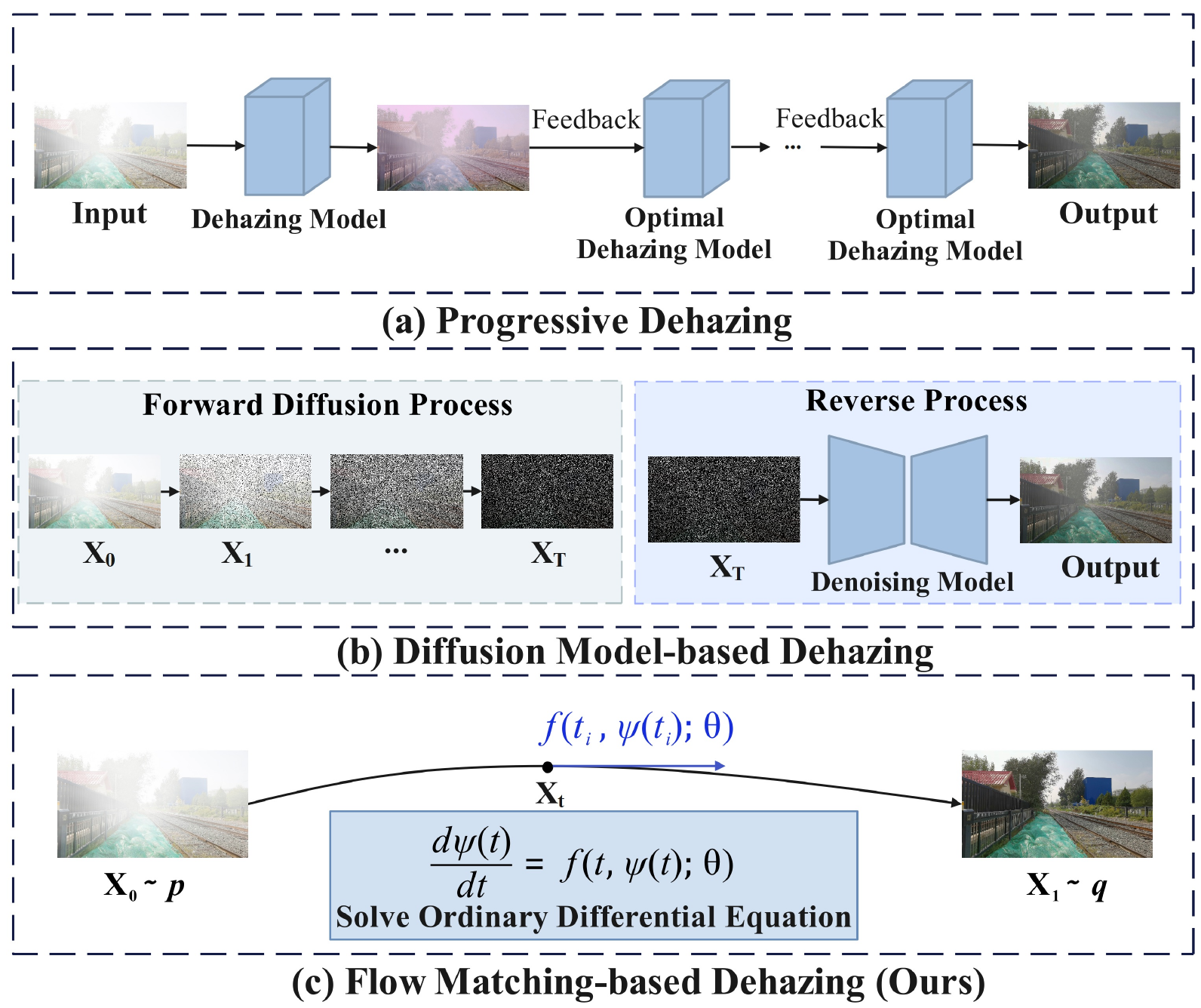} 
	\caption{Comparison of different dehazing paradigms. 
		(a) Progressive dehazing performs restoration step-by-step, which may lead to suboptimal convergence at the local optima. 
		(b) Diffusion model-based dehazing achieves global optimization through iterative sampling, but has limitations including high computational cost and high noise sensitivity. 
		(c) Our Flow Matching-based dehazing method models the dehazing process as an ordinary differential equation of \textbf{continuous} vector fields, enabling efficient and stable dehazing with adaptive color correction.
	}
	\label{figure1}
	\vspace{-5mm}
\end{figure}

%
Existing UHD dehazing methods, such as 4KDehazing~\cite{Zheng_2021_CVPR}, 4K-Haze~\cite{Su_2023_SSRN}, and LapDehazeNet~\cite{LapDehazeNet}, rely on large convolutional networks, resulting in increased computational complexity. 
For example, 4KDehazing~\cite{Zheng_2021_CVPR} has 34.55M parameters, which significantly increases computational complexity and prolongs inference time. 
Lightweight UHD restoration models like UHDformer~\cite{UHDformer} and UHDDIP~\cite{UHDDIP} have smaller parameter sizes (0.34M and 0.81M, respectively), relying on additional priors such as normal maps, which limits their dehazing performance compared to task-specific models. 
%
To solve these challenges, we propose a Flow Matching method with a dehazing prior, which uses a differential equation to continuously and gradually optimize the vector field to find the optimal solution for dehazing. 
%
%
%
%
Compared to UHDformer~\cite{UHDformer} and UHDDIP~\cite{UHDDIP}, our method boosts PSNR by approximately 2dB, achieving a better trade-off between performance and efficiency. 
%

It is important to clarify that dehazing based on Flow Matching differs from progressive and diffusion model-based methods. 
%
As shown in Figure~\ref{figure1}, \mbox{\textbf{Progressive Dehazing}} methods \cite{LAP-Net_cvpr19,li2023pfonet,cheng2024progressive,liang2021progressive} restore images through stage-wise optimization, using multi-scale features and iterative refinement to mitigate detail loss and over-dehazing. 
%
However, discrete iterations and fixed steps often lead to suboptimal convergence at the local optima and color shifts in complex haze conditions. 
%
%
%
%
%
%
\mbox{\textbf{Diffusion Model-based Dehazing}} methods \cite{liu2024diff,yu2023high,cheng2024dehazediff} optimize globally by simulating data diffusion but require hundreds of iterations, resulting in long inference times and noise-induced artifacts in low-texture regions. 
Despite accelerated sampling strategies \cite{yin2024one,liu2022flow,li2025one}, diffusion models still struggle to balance efficiency and fidelity.

%
%
%
%
We propose a UHD image dehazing method, \mbox{\textbf{4KDehazeFlow}}, which designs a dedicated, smooth, and interpretable dehazing path based on Flow Matching. 
%
Specifically, we build a general Flow Matching dehazing method that integrates the initial and target distributions, vector field, and ODE solver, enabling effective UHD image restoration under limited computational resources. 
Within this framework, at first, the hazy images are defined as samples drawn from the initial distribution, while their corresponding clear counterparts constitute the target distribution. The goal is to gradually map the initial distribution to the target distribution through Flow Matching. 
%
Second, during the Flow Matching process, the vector field guides the transformation path for image dehazing from the hazy images to their clean counterpart. To optimize the dehazing effect, our improved CNN network includes a haze purifier and a dehazing prior, and is further improved with Haze-LUT for adaptive colour and brightness adjustment, which enhances dehazing accuracy and visual quality. 
Finally, we employ a fourth-order Ronger-Kuta (RK4) differential equation solver, which stably solves the dehazing flow field through precise step-by-step iterations. The RK4 method effectively balances computational accuracy and stability by calculating a weighted average of four stages at each time step, ensuring dehazing finesse and reliability. 
%

Our main contributions are summarized as follows:
\begin{itemize}
	\item We propose 4KDehazeFlow, which overcomes the local convergence issues in traditional discrete iterations. Furthermore, the vector field is modeled as an atmospheric scattering purifier, which enables a more accurate and effective adaptation to the image dehazing task.
	
	\item We propose a learnable Haze-LUT module that compresses complex haze transformation parameters into a compact 3D mapping matrix. This module incorporates an adaptive nonlinear color transformation mechanism to enhance both dehazing accuracy and visual quality.
	
	\item Extensive qualitative and quantitative experiments demonstrate that, on 4K dehazing datasets, our method outperforms existing methods across multiple evaluation metrics, achieving state-of-the-art results in both performance and efficiency.
	
\end{itemize}

\begin{figure*}[t]
	\centering
	\includegraphics[width=2.08\columnwidth]{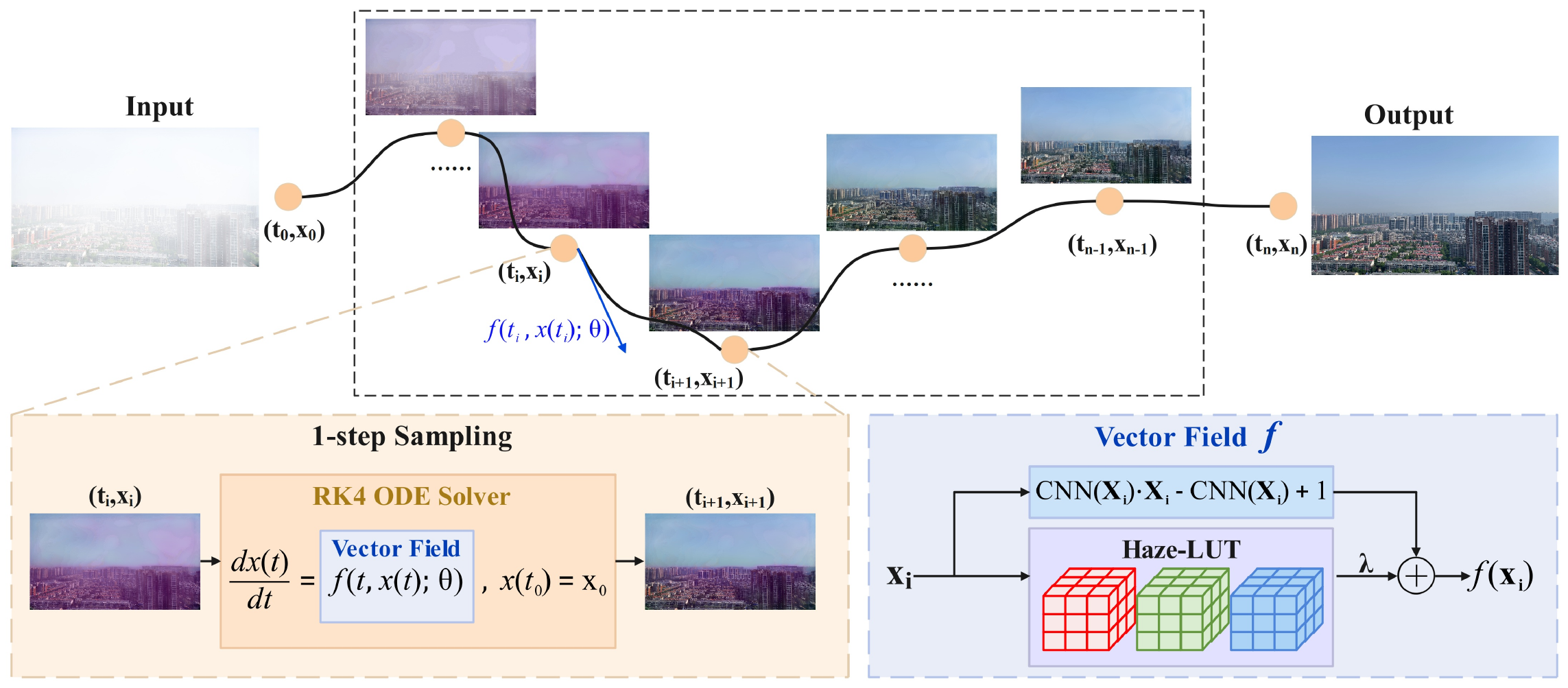} 
	\caption{The overall architecture of 4KDehazeFlow. Our method employs a continuous stepwise inference process with adjustable amount of inference steps. 
		The core innovation lies in the vector field, which integrates two key components: (1) an atmospheric scattering purifier implemented through the CNN feature extraction with potential dehazing function, and (2) a data-driven learned 3D LUT for degraded color reconstruction.
		Notably, the role of ODE is to optimize the entire vector field.
	}
	\vspace{-4mm}
	\label{UHDehazeFlow}
\end{figure*}

\section{Related Works}
\noindent \textbf{Single Image Dehazing.}
Currently, deep learning-based methods overcome traditional dehazing models' limitations by learning end-to-end models that adaptively extract features from data. 
%
Existing UHD dehazing methods, such as 4KDehazing~\cite{Zheng_2021_CVPR}, 4K-Haze~\cite{Su_2023_SSRN}, and LapDehazeNet~\cite{LapDehazeNet}, rely on traditional convolutional neural networks with large model parameters. 
%
%
For example, the 4KDehazing model contains 34.55 million trainable parameters, leading to excessive computational complexity and prolonged inference time. 
In contrast, existing UHD image restoration frameworks such as UHDformer~\cite{UHDformer} and UHDDIP~\cite{UHDDIP} exhibit more compact architectures, with only 0.34 million and 0.81 million parameters, respectively. However, these methods rely on supplementary prior information (e.g., normal maps) and demonstrate suboptimal dehazing performance compared to task-specific models. 
Meanwhile, non-UHD dehazing methods \cite{zhang2024depth, PTTD_eccv24, chen2024dea, Mb-Taylorformer_ICCV23, C2PNet_cvpr23, Dehamer_cvpr22, DehazeFlow_MM21} have made significant progress for lower-resolution images dehazing. 
However, when applied to UHD images, they often encounter memory overflow or suboptimal performance due to the increased computational demands and difficulty in preserving fine details. 
Therefore, it is crucial to develop methods that are effective for both UHD and non-UHD image dehazing.

\noindent \textbf{Flow Matching.}
Flow Matching \cite{lipman2022flow} is a simple yet powerful framework for generative modeling that has significantly pushed the boundaries of the state-of-the-art across various domains and large-scale applications, including generation of images~\cite{esser2024scaling,hu2024latent}, audios~\cite{vyas2023audiobox,yun2025flowhigh}, videos~\cite{cao2025video}, speech~\cite{le2024voicebox}, and point clouds~\cite{liu2025efficient}. 
It enables simulation-free training of Continuous Normalizing Flow (CNF), offering enhanced efficiency compared to traditional diffusion training and sampling methods. 
%
%
UniFlowRestore~\cite{UniFlowRestore} models video restoration as the continuous evolution of a physical vector field through Flow Matching and prompt guidance, enabling unified restoration across multiple tasks. 
Building on Flow Matching's success across various domains, we apply this framework to tackle the unique challenges in UHD image dehazing. 
By directly learning the mapping between data distributions, this approach avoids the complex iterative processes characteristic of traditional generative models, significantly reducing both computational complexity and memory usage. 
Our goal is to strike an optimal balance between performance and efficiency in UHD image dehazing through Flow Matching.

\noindent \textbf{3D LUTs for Image Enhancement.}
3D LUTs are widely used in camera imaging pipelines and photo editing software for efficient color mapping.  
Recently, learnable LUT methods have been explored for image enhancement \cite{li2024real,conde2024nilut,LUT_tip23,Yang_2022_CVPR,LUT_TPAMI22,Wang_2021_ICCV}. 
We employ 3D LUT to adaptively adjust color transformations during dehazing, enhancing visual quality while preserving natural color restoration.

\section{Methodology}
\noindent \textbf{Overview\\}
An overview of 4KDehazeFlow is shown in Figure~\ref{UHDehazeFlow}. 
Given a haze input image $\mathbf{X_0}$, 4KDehazeFlow gradually recovers the clear image by solving a time-dependent Ordinary Differential Equation (ODE), defined by a haze-aware vector field $f$, which models the flow of samples from the hazy distribution $p_0$ to the target distribution $p_1=q$.
In each step, the input image $\mathbf{X_i}$ is fed into the haze-aware vector field $f$, which incorporates an atmospheric scattering purifier to clean up the haze, along with a Haze-LUT to adaptively adjust the image's color. 
Then the RK4 ODE solver is used to update the image $\mathbf{X_{i+1}}$ at each time step based on the haze-aware vector field. 
%
This process iterates through all time steps, ultimately yielding the clear output image $\mathbf{X_n}$.

\subsection{Flow Matching for 4KDehazeFlow}
Our work presents a Flow Matching-based dehazing method, termed 4KDehazeFlow. The proposed method consists of four key components: the initial distribution, target distribution, vector field, and ODE solver. 
%
In 4KDehazeFlow, hazy images are defined as the initial distribution $p$, while the target distribution $q$ is defined by the space of clear  images. 
The objective is to gradually transform hazy images into clear images by modeling the haze removal process through a time-dependent vector field $f_t$. 
%
The vector field $f_t$ is composed of the atmospheric scatter purifier and the Haze-LUT, which extract multi-scale features and perform adaptive color correction. 
The dehazing process is governed by the following Ordinary Differential Equation (ODE):
\begin{align}
	&\frac{d x(t)}{dt}=f(t,x(t);\theta)\     ,\ x(\mathbf{t_0})=\mathbf{X_0},
\end{align}
where $dt$ denotes the time step, $x(t)$ is the image at time $t$, $f(t,x(t);\theta)$ is the haze-aware vector field parameterized by $\theta$, and $t_0$ is the initial time step. 
By solving this ODE, the image is progressively guided along the probability path from the hazy state to the clear state.

\subsection{RK4 ODE Solver}
To solve Equation (1), we employ the Runge-Kutta 4th order (RK4) method. 
RK4 is a widely used numerical technique for approximating solutions to differential equations with high accuracy \cite{butcher2016numerical}. 
In 4KDehazeFlow, the RK4 solver iteratively updates the hazy image at each time step, ensuring a smooth and stable transition from the initial hazy state to the clear image. 
At each step, the RK4 solver computes intermediate slopes to approximate the solution of the ODE. The update rule for the image $\mathbf{X_{i+1}}$ at time $t_{i+1}$, based on the current image $\mathbf{X_{i}}$ at time $t_i$ and the vector field $f_t$, is expressed as:
\begin{align}
	&\mathbf{X_{i+1}}=\mathbf{X_i} + \frac{1}{6}(k_1+2k_2+2k_3+k_4)dt,
\end{align}
where the intermediate slopes $k_1$, $k_2$, $k_3$, and $k_4$ are computed as follows:
\begin{align}
	k_1 &= f(t_i, \mathbf{X_i}),\\
	k_2 &= f(t_i+\frac{dt}{2}, \mathbf{X_i}+\frac{k_1}{2}\cdot dt),\\
	k_3 &= f(t_i+\frac{dt}{2}, \mathbf{X_i}+\frac{k_2}{2}\cdot dt),\\
	k_4 &= f(t_i+dt, \mathbf{X_i}+k_3\cdot dt).
\end{align}
After computing these values, the image at the next time step $t_{i+1}=t_i+dt$ is updated to $\mathbf{X_{i+1}}$.
\subsection{Haze-Aware Vector Field}
\label{sec:Vector_Field}
During the Flow Matching process, the vector field specifies the transformation path for image dehazing, modeling the dynamic change of the image distribution from hazy to clear over time. 
To achieve this, we develop an atmospheric scattering purifier inspired by the AODNet~\cite{AODNet}, which directly generates the clear image through a lightweight CNN. This approach captures multi-scale features and improves dehazing by avoiding separate transmission and atmospheric light estimation. 
%
However, since the Flow Matching can run the function of dehazing in multiple steps, it can result in a degradation of color fidelity. 
To further enhance color accuracy, we propose Haze-LUT, a 3D lookup table (LUT) that adaptively adjusts the image's color through a nonlinear color transformation mechanism. 
This process can be written as:
\begin{equation}
	f(\mathbf{X_i})=\mathbf{O_m}+\lambda \cdot \mathbf{O_{lut}},
\end{equation}
where $f(\cdot)$ represents the output of haze-aware vector field, $\mathbf{X_i}$ is the current input image, $\mathbf{O_m}$ is the output image obtained from the atmospheric scattering purifier, $\mathbf{O_{lut}}$ is the color adjustment derived from the Haze-LUT, and $\lambda$ is a scaling factor that controls the strength of the color transformation.

\noindent \textbf{Atmospheric Scattering Purifier.}
The atmospheric scattering purifier in 4KDehazingFlow plays a crucial role in extracting multi-scale features from the hazy input image to generate a clearer output. 
The CNN directly generate the clear image without explicitly estimating the transmission map or atmospheric light, offering a lightweight and efficient dehazing solution. 
%
%
%
AODNet reformulates the atmospheric scattering model as:
\begin{align}
	&\mathrm{J(x) = K(x) \cdot I(x) - K(x) +b}, 
\end{align}
where $\mathrm{J(x)}$ is the clear image, $\mathrm{I(x)}$ is the input hazy image, $\mathrm{K(x)}$ is a learnable scaling parameter and $\mathrm{b}$ is a learnable bias term initialized to 1.

Following this idea, our CNN is designed to estimate $\mathrm{K(x)}$ directly from the input, and the dehazed image is obtained as:
%
\begin{align}
	&\mathbf{O_m}=\mathrm{CNN(\mathbf{X_i}) \cdot \mathbf{X_i} - CNN(\mathbf{X_i})+1},
\end{align}
where $\mathbf{X_i}$ is current input image, $\mathbf{O_m}$ is the output of enhanced CNN.

The CNN network consists of three down-sampling encoder blocks, a spatial attention module, and three up-sampling decoder blocks:
\begin{align}
	&\mathrm{CNN(\mathbf{X_i})}=\mathrm{D(Attn(E(\mathbf{X_i})))+\mathbf{X_i}},
\end{align}
where $\mathbf{X_i}$ is the input image, $\mathrm{E(\cdot)}$ denotes the down-sampling encoder blocks, $\mathrm{Attn(\cdot)}$ refers to the attention mechanism, and $\mathrm{D(\cdot)}$ represents the up-sampling decoder blocks. 
%
%
%
The encoder progressively reduces spatial dimensions while extracting hierarchical features through max-pooling, convolution, adaptive normalization, and GELU activations. The spatial attention module highlights important regions, while the decoder refines features via bilinear up-sampling and skip connections to restore fine details. 
This network uses convolution kernels of size $1\times1$ or $3\times3$.
\begin{table*}[t]
	\centering
	\caption{Quantitative comparison on 4KID~\cite{Zheng_2021_CVPR}, I-Haze~\cite{I-HAZE}, and O-Haze~\cite{O-HAZE} datasets. The best and second-best values are highlighted with \textbf{bold} text and \underline{underlined} text, respectively.}
	\resizebox{\textwidth}{!}{
		\begin{tabular}{cl|ccc|ccc|ccc|ccc|c}
			\toprule
			\multicolumn{2}{c|}{\multirow{2}{*}{\textbf{Methods}}} & \multicolumn{3}{c|}{\textbf{4KID}~\cite{Zheng_2021_CVPR}} & \multicolumn{3}{c|}{\textbf{I-Haze}~\cite{I-HAZE}} & \multicolumn{3}{c|}{\textbf{O-Haze}~\cite{O-HAZE}}&\multicolumn{3}{c|}{\textbf{Average}}&\multicolumn{1}{c}{\textbf{Overhead}} \\
			\cline{3-15}
			& & \textbf{PSNR}$\boldsymbol{\uparrow}$ & \textbf{SSIM}$\boldsymbol{\uparrow}$ & \textbf{LPIPS}$\boldsymbol{\downarrow}$ & \textbf{PSNR}$\boldsymbol{\uparrow}$ & \textbf{SSIM}$\boldsymbol{\uparrow}$ & \textbf{LPIPS}$\boldsymbol{\downarrow}$ & \textbf{PSNR}$\boldsymbol{\uparrow}$ & \textbf{SSIM}$\boldsymbol{\uparrow}$ & \textbf{LPIPS}$\boldsymbol{\downarrow}$ &\textbf{PSNR}$\boldsymbol{\uparrow}$ & \textbf{SSIM}$\boldsymbol{\uparrow}$ & \textbf{LPIPS}$\boldsymbol{\downarrow}$ &\textbf{Param}$\boldsymbol{\downarrow}$\\
			\hline
			\multirow{4}{*}{\textbf{non-UHD}} & Dehamer~\cite{Dehamer_cvpr22}& 12.56 & 0.7383 & 0.2784 & 14.91 & 0.7469 & 0.4430 & 15.27 & 0.4256 & 0.5396 &14.25 &0.6369 &0.4203 &132.50M\\
			& C2PNet~\cite{C2PNet_cvpr23}& 15.61 & 0.7642 & 0.3000 & 16.65 & 0.7383 & \underline{0.4174} & 16.66 & 0.6454 & 0.4595& 16.30& 0.7160& 0.3923&7.17M \\
			& MB-TaylorFormer~\cite{Mb-Taylorformer_ICCV23} & 18.92 & 0.8054 & 0.3010 & 17.01 & 0.6634 & 0.4365 & \textbf{22.98} & \underline{0.7030} & 0.4773&19.64 & 0.7239& 0.4090&7.43M \\
			& PTTD~\cite{PTTD_eccv24}& 18.31 & 0.7738 & 0.4122 & 15.02 & 0.6761 & 0.4994 & 19.43 & 0.7018 & 0.4521&17.59 &0.7172 &0.4546 &2.61M \\
			\hline
			\multirow{4}{*}{\textbf{UHD}} & 4KDehazing~\cite{Zheng_2021_CVPR}& 20.71 & 0.8592 & 0.2948 & 16.76 & 0.7633 & 0.4620 & 18.43 & 0.6912 & \underline{0.4396} &18.63 &0.7712 &0.3988 &34.55M\\
			& UHDformer~\cite{UHDformer}& 22.59 & 0.9327 & 0.1588 & \underline{18.80} & 0.7668 & 0.4966 & 16.27 & 0.5971 & 0.4547&19.22 &0.7655 & 0.3700&\textbf{0.34M} \\
			& UHDDIP~\cite{UHDDIP}& \underline{23.74} & \textbf{0.9421} & \underline{0.1221} & 17.27 & \underline{0.7699} & 0.4592 & 17.61 & 0.6383 &0.4864& \underline{19.54}& \underline{0.7834}& \underline{0.3152} &\underline{0.81M} \\
			& \textbf{4KDehazeFlow (Ours)}& \textbf{23.85} & \underline{0.9379} &\textbf{0.1101}  & \textbf{19.63} & \textbf{0.8085} & \textbf{0.3881} & \underline{21.39} & \textbf{0.8138} & \textbf{0.4391} &\textbf{21.62} &\textbf{0.8534} &\textbf{0.3124} &7.90M\\
			\bottomrule
	\end{tabular}	}
	\vspace{-4mm}
	\label{table-UHD}
\end{table*}

\begin{figure*}[t]
	\centering
	\includegraphics[width=2.1\columnwidth]{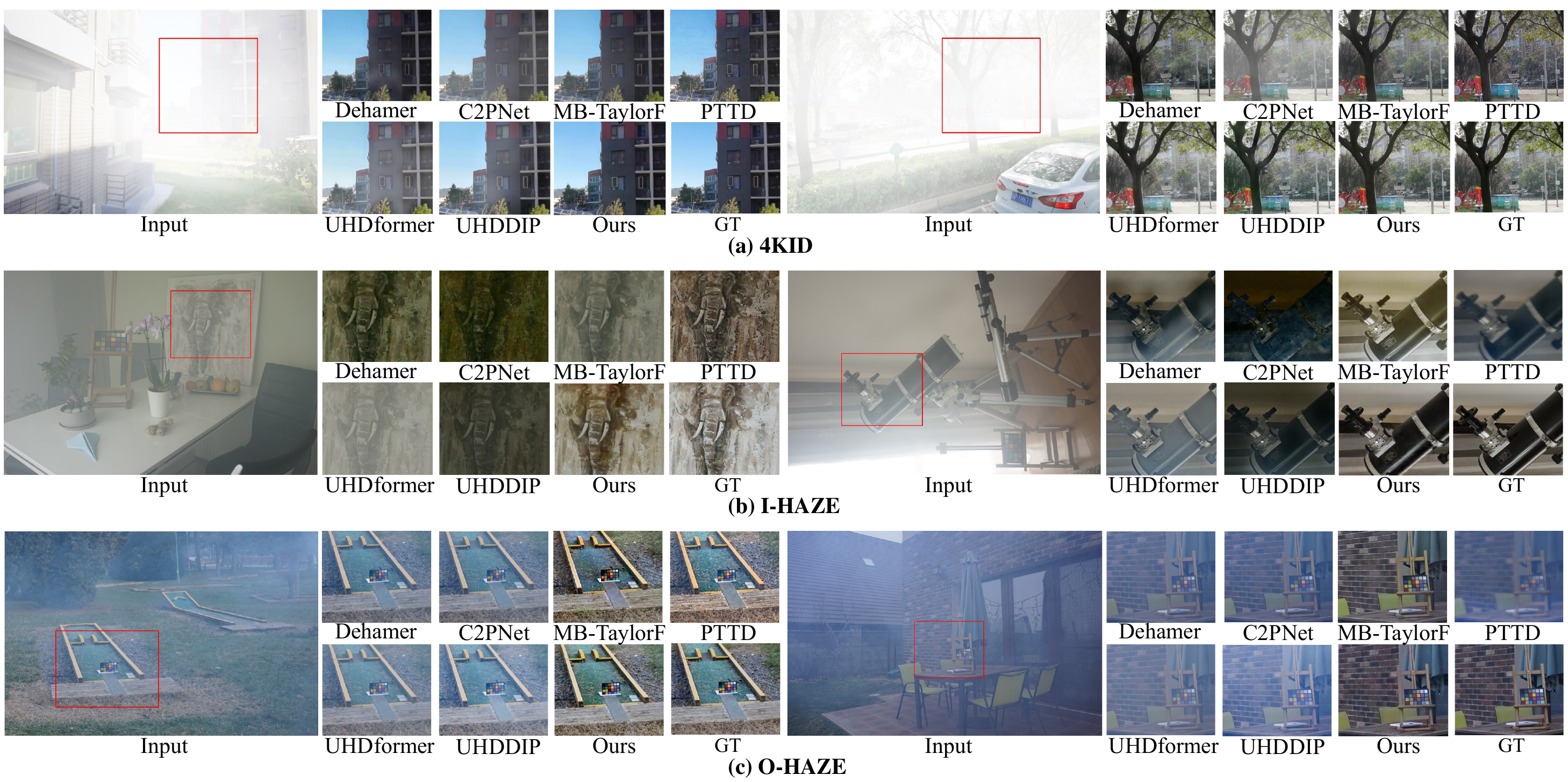} 
	\vspace{-3mm}
	\caption{Visual comparison on 4KID~\cite{Zheng_2021_CVPR}, I-HAZE~\cite{I-HAZE}, and O-HAZE~\cite{O-HAZE} datasets. 
		%
		The results on UHD (4K) scenes demonstrate that the proposed 4KDehazeFlow consistently outperforms state-of-the-art methods across varying haze densities and diverse scene complexities.
	}
	\vspace{-5mm}
	\label{figure-UHD}
\end{figure*}

\noindent \textbf{Haze-LUT.}
While the atmospheric scattering purifier effectively removes haze, it often results in color loss or distortions in the process. 
To mitigate these issues, we propose Haze-LUT, a 3D LUT that adaptively adjusts the image's color through a nonlinear color transformation mechanism. 
A 3D LUT represents a three-dimensional lattice consisting of $\mathrm{M^3}$ elements, denoted as $\mathrm{\{ H_{(i,j,k)}\}}_{\mathrm{i,j,k=0,1,...,M-1}}$, where $\mathrm{M}$ is the number of bins per color channel. 
Each element $\mathrm{H_{(i,j,k)}}$ corresponds to the indexing RGB color $\mathrm{\{ r_{(i,j,k)}^I,g_{(i,j,k)}^I,b_{(i,j,k)}^I \}}$ and its transformed output RGB color $\mathrm{\{ r_{(i,j,k)}^O,g_{(i,j,k)}^O,b_{(i,j,k)}^O \}}$. 
The precision of this transformation is determined by $\mathrm{M}$, which is typically set to 33 in practice \cite{zeng2020learning}. 
The Haze-LUT transformation is performed in two key steps: lookup and trilinear interpolation. 

For an input RGB color $\mathrm{\{ r_{(x,y,z)}^I,g_{(x,y,z)}^I,b_{(x,y,z)}^I \}}$, its lattice coordinates $\mathrm{(x',y',z')}$ are computed as:
\begin{align}
	&\mathrm{x'=\frac{r_{(x,y,z)}^I}{s}, y'=\frac{g_{(x,y,z)}^I}{s}, z'=\frac{b_{(x,y,z)}^I}{s}},
	&\mathrm{s=\frac{C_{max}}{M}}
\end{align}
where $C_{max}$ is the maximum color value. 
Once the location is determined, the output RGB color $\mathrm{\{ r_{(x',y',z')}^O,g_{(x',y',z)}^O,b_{(x',y',z')}^O \}}$ is obtained through trilinear interpolation from the 8 adjacent lattice vertices. 
Finally, the Haze-LUT correction is expressed as:
\begin{align}
	&\mathbf{O_{lut}}=\mathrm{IP(\mathbf{X_i},H)},
\end{align}
where $\mathbf{X_i}$ represents the input image, $\mathrm{IP(\cdot)}$ denotes the trilinear interpolation function, and $\mathrm{H}$ is the Haze-LUT containing the sampled output RGB values.

\subsection{Loss Function}
To optimize the weights and biases of the network, we utilize the L1 loss in the RGB color space as the fundamental reconstruction loss. This approach ensures that the network accurately captures and reproduces the fine details and color information of the input images. L1 loss can be written as:
\begin{align}
	&\mathrm{L = \lVert \mathbf{Y} - \mathbf{\hat{Y}} \rVert_1},
\end{align}
where $\mathbf{Y}$ denotes the ground truth, and $\mathbf{\hat{Y}=X_n}$ represents the final output.

\begin{figure*}[t]
	\centering
	\includegraphics[width=2.1\columnwidth]{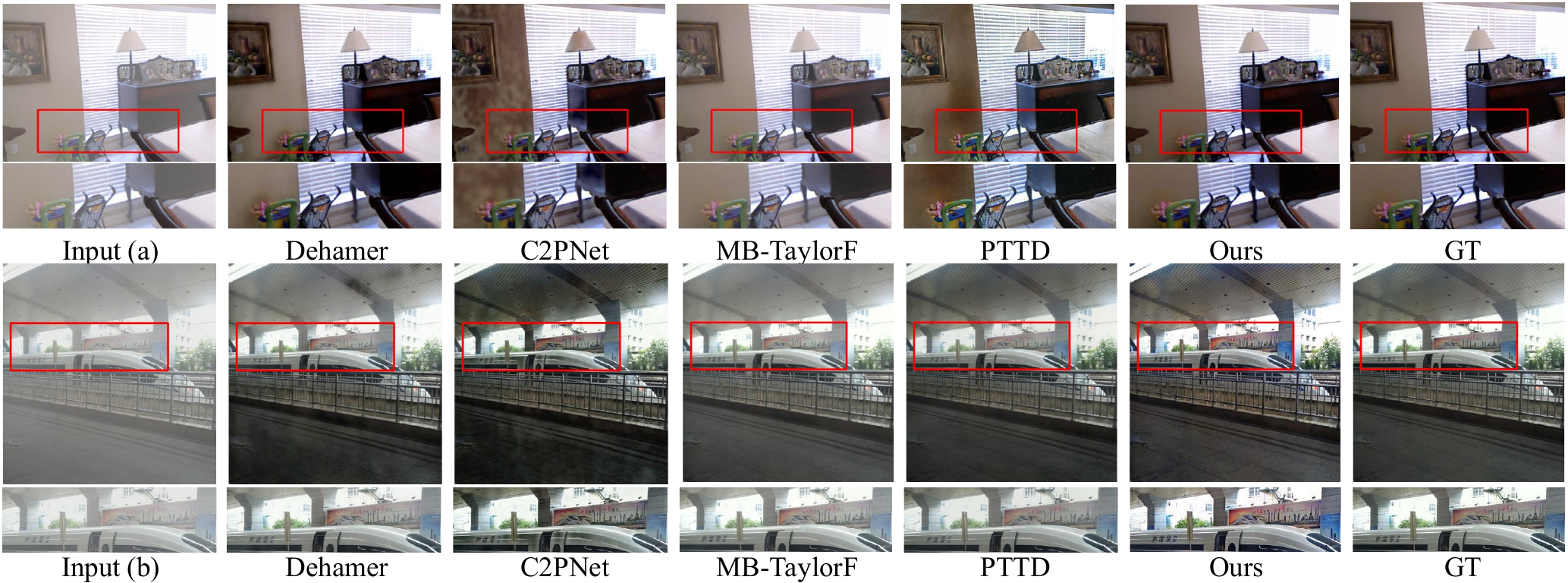} 
	\caption{Visual comparison on the SOTS~\cite{SOTS} dataset.
		The results highlight that the proposed 4KDehazeFlow achieves superior image restoration performance in low-resolution scenes.
	}
	\vspace{-4mm}
	\label{figure-SOTS}
\end{figure*}

\section{Experiments}
In this section, we evaluate the proposed method through comprehensive experiments on both UHD and non-UHD datasets. 
We compare our approach with seven state-of-the-art image dehazing methods: Dehamer~\cite{Dehamer_cvpr22}, C2PNet~\cite{C2PNet_cvpr23}, MB-TaylorFormer~\cite{Mb-Taylorformer_ICCV23}, PTTD~\cite{PTTD_eccv24}, 4KDehazing~\cite{Zheng_2021_CVPR}, UHDformer~\cite{UHDformer}, and UHDDIP~\cite{UHDDIP}. 
Additionally, we perform ablation studies to evaluate the contribution and effectiveness of each module in our method.
More experimental results are provided in the Supplementary Materials.

\begin{table}[t]
	\centering
	\footnotesize
	
	\caption{Computational efficiency comparison of UHD  methods. The best and second-best values are highlighted with \textbf{bold} text and \underline{underlined} text, respectively.}
	\vspace{-3mm}
	\resizebox{0.47\textwidth}{!}{
		\begin{tabular}{l|c c c}
			\toprule
			\textbf{Methods} & \textbf{TIME$\boldsymbol{\downarrow}$} & \textbf{MACs$\boldsymbol{\downarrow}$} & \textbf{Param$\boldsymbol{\downarrow}$}\\
			\hline
			4KDehazing~\cite{Zheng_2021_CVPR}&\textbf{0.0948s}&\underline{105.90G} &34.55M \\
			UHDformer~\cite{UHDformer}&0.4273s &382.64G  &\textbf{0.34M} \\
			UHDDIP~\cite{UHDDIP}&0.4280s&274.72G  &\underline{0.81M} \\
			\textbf{4KDehazeFlow (Ours)}&\underline{0.1540s}  &\textbf{40.28G } & 7.90M\\
			\bottomrule
		\end{tabular}
	}
	\label{table-overhead}
	\vspace{-5mm}
\end{table}
\subsection{Datasets}
%
%
%
%
\noindent \textbf{UHD datasets.} We evaluate on three UHD dehazing datasets: 4KID~\cite{Zheng_2021_CVPR}, I-HAZE~\cite{I-HAZE}, and O-HAZE~\cite{O-HAZE}. 4KID~\cite{Zheng_2021_CVPR} contains 3480$\times$2160 images, while I-HAZE and O-HAZE offer high-quality indoor/outdoor hazy-clear pairs. We select 4K+ resolution images, using 70\% for training and 30\% for testing. 

\noindent \textbf{non-UHD dataset.} 
The SOTS~\cite{SOTS} dataset, consisting of both indoor and outdoor hazy-clear image pairs with a resolution of 608 $\times$ 448, is mixed together in our experiments. We randomly select 700 images for training and 300 images for testing.

\begin{figure}[t]
	\centering
	\includegraphics[width=\columnwidth]{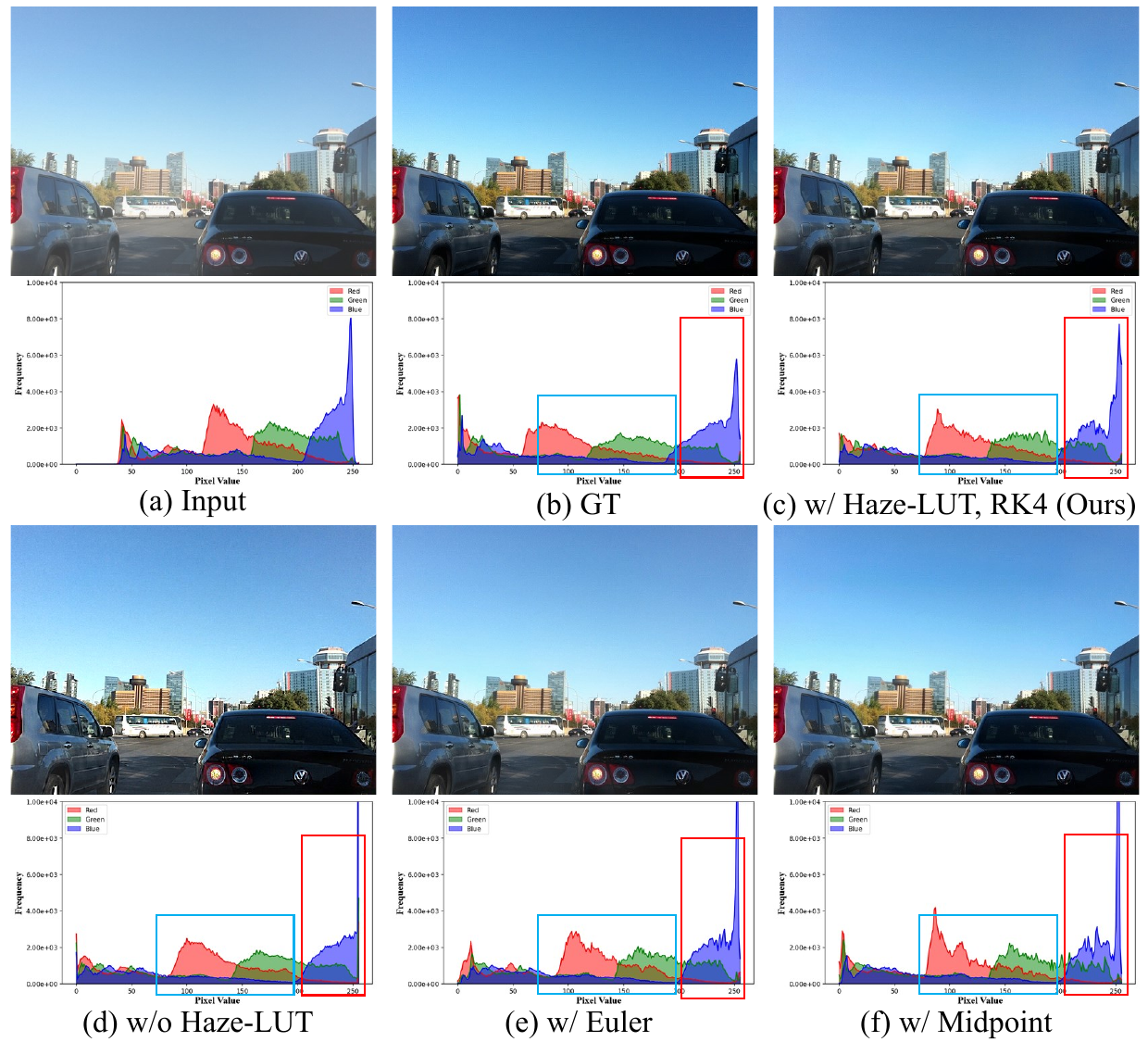} 
	
	\caption{Visualization results of ablation experiments. 
		%
		As evidently demonstrated, the removal of key components leads to significant degradation, including a higher blue peak and greater red deviation from the ground truth values.
	}
	
	\label{figure-ab}
	\vspace{-5mm}
\end{figure}
\subsection{Implementation Details}
%
%
%
%
%
Experiments are conducted using PyTorch on a single NVIDIA A100 80GB GPU. We adopt the AdamW optimizer (weight decay $1\times10^{-4}$). The initial learning rates and batch sizes are: 0.001/6 for 4KID~\cite{Zheng_2021_CVPR}, 0.0001/4 for I-HAZE~\cite{I-HAZE} and O-HAZE~\cite{O-HAZE}, and 0.001/8 for SOTS~\cite{SOTS}. The ReduceLROnPlateau scheduler lowers the learning rate by 0.5 if validation loss stagnates for 100 epochs. 
We evaluate performance using PSNR, SSIM~\cite{psnr_ssim}, LPIPS~\cite{lpips}, and NIQE~\cite{NIQE}.
\begin{table}[t]
	\centering
	\small
	\caption{Comparison of quantitative results on SOTS~\cite{SOTS} dataset. The best and second-best values are highlighted with \textbf{bold} text and \underline{underlined} text, respectively.}
	\vspace{-3mm}
	\resizebox{0.47\textwidth}{!}{
		\begin{tabular}{l|c c c}
			\toprule
			\textbf{Methods} & \textbf{NIQE$\boldsymbol{\downarrow}$} & \textbf{SSIM$\boldsymbol{\uparrow}$} & \textbf{LPIPS$\boldsymbol{\downarrow}$}\\
			\hline
			Dehamer~\cite{Dehamer_cvpr22}&3.82  &0.9700 &\underline{0.0530}  \\
			C2PNet~\cite{C2PNet_cvpr23}&3.73  & \underline{0.9806} &0.0751 \\
			MB-TaylorFormer~\cite{Mb-Taylorformer_ICCV23}&4.04 &\textbf{0.9810} &0.0648\\
			PTTD~\cite{PTTD_eccv24}&\underline{3.62} &0.8228 &0.1279 \\
			4KDehazing~\cite{Zheng_2021_CVPR}&3.65&0.8366 &0.1242 \\
			UHDformer~\cite{UHDformer}&4.81 &0.8249  &0.1563 \\
			UHDDIP~\cite{UHDDIP}&4.97&0.8496  &0.1420 \\
			\textbf{4KDehazeFlow (Ours)}&\textbf{3.47}  &0.9709  & \textbf{0.0478}\\
			\bottomrule
		\end{tabular}
	}
	\label{table-SOTS}
	\vspace{-3mm}
\end{table}
\begin{table}[t]
	\centering
	\caption{Ablation study of the key components in 4KDehazeFlow.}
	\vspace{-3mm}
	\resizebox{0.47\textwidth}{!}{
		\begin{tabular}{c|c|c|ccc}
			\toprule
			\textbf{Ablation} & \textbf{Datasets} & \textbf{Setting} & \textbf{PSNR}$\boldsymbol{\uparrow}$ & \textbf{SSIM}$\boldsymbol{\uparrow}$ & \textbf{LPIPS}$\boldsymbol{\downarrow}$ \\
			\hline
			\multirow{3}{*}{\textbf{Haze-LUT}} & \multirow{3}{*}{SOTS} &$\times$  &27.75  & 0.9107 &0.1054  \\
			&      & Fixed &24.70  &0.7698  &0.3091  \\
			&      & \textbf{Learnable} &\textbf{33.38}  &\textbf{0.9709}  &\textbf{0.0478}  \\
			\hline
			\multirow{3}{*}{$\boldsymbol{\lambda}$} & \multirow{3}{*}{4KID} & 0.1 &22.59  &0.8937  &0.2041  \\
			&       & \textbf{0.5} &\textbf{23.85}  &\textbf{ 0.9379} & \textbf{0.1101} \\
			&       & 1 &21.68  &0.8858  & 0.1840 \\
			\hline
			\multirow{3}{*}{\textbf{ODE Solver}} & \multirow{3}{*}{SOTS} & Euler & 30.12 & 0.9518 & 0.0509 \\
			&  &Midpoint  &32.82  & 0.9671 & 0.0553 \\
			&  &\textbf{RK4}  &\textbf{33.38} & \textbf{0.9709} &\textbf{0.0478} \\
			\bottomrule
		\end{tabular}
	}
	\label{table-ab}
	\vspace{-5mm}
\end{table}
\subsection{Quantitative Results}
%
%
%
\noindent \textbf{Quantitative performance metrics.}
Table~\ref{table-UHD} summarizes the comparison with SOTA methods. On 4KID~\cite{Zheng_2021_CVPR}, 4KDehazeFlow achieves the highest PSNR 23.85dB and the lowest LPIPS 0.1101, surpassing UHDDIP~\cite{UHDDIP} by 0.11dB in PSNR and 0.012 in LPIPS. 
On I-HAZE~\cite{I-HAZE}, it achieves PSNR 19.63dB, 0.83dB higher than UHDformer; SSIM 0.8085, 0.0386 higher than UHDDIP; and LPIPS 0.3881, 0.0293 lower than C2PNet. 
On O-HAZE~\cite{O-HAZE}, our method reaches SSIM 0.8138, outperforming MB-TaylorF by 0.1108, and achieves LPIPS 0.4391. 
Across the three datasets, 4KDehazeFlow achieves the highest average PSNR 21.62dB, 2.08dB above UHDDIP; SSIM 0.8534, 0.07 higher than UHDDIP; and the lowest LPIPS 0.3124, indicating strong restoration and perceptual quality. 
On the SOTS~\cite{SOTS} dataset, Table~\ref{table-SOTS} shows that 4KDehazeFlow remains competitive on non-UHD images.

\noindent \textbf{Complexity of the model.}
Table~\ref{table-overhead} compares the computational overhead of UHD dehazing methods. 
Compared to UHDformer and UHDDIP, 4KDehazeFlow achieves significantly lower MACs with competitive efficiency. 
Note that our method can control the number of steps in Flow Matching to regulate the inference time.

\begin{figure*}[t]
	\centering
	\includegraphics[width=2.1\columnwidth]{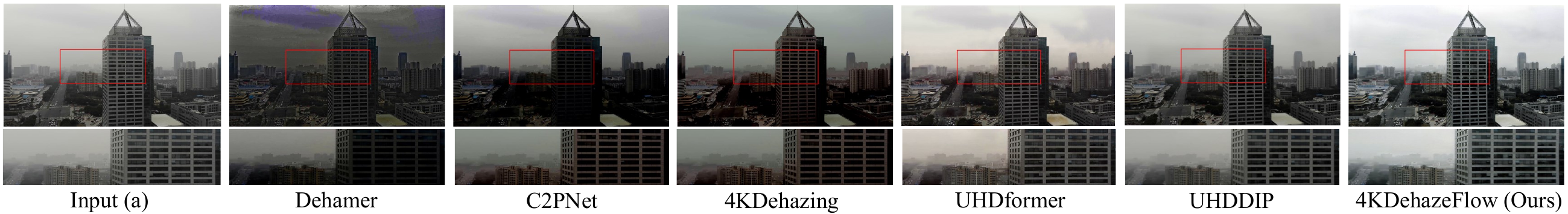} 
	\caption{Visualization of real-world 4K hazy images and dehazed results. 
		Compared to other existing methods, the dehazed images generated by our approach exhibit superior clarity, effectively demonstrating its robustness in handling real-world haze scenarios.
	}
	\label{figure-real}
	\vspace{-5mm}
\end{figure*}

%
%
%
%

%

\subsection{Qualitative Results}
We provide visual comparisons with SOTA methods to further demonstrate the effectiveness of our approach. 
In Figure~\ref{figure-UHD}(a), under heavy haze, our method restores cleaner structures and more natural colors, avoiding the whitish residues and sky artifacts seen in other methods. 
Figure~\ref{figure-UHD}(b) shows superior texture recovery in indoor UHD scenes, while Figure~\ref{figure-UHD}(c) highlights our advantage in handling haze and preserving detail in 4K images. 
Figure~\ref{figure-SOTS} shows that on low-resolution images, our method strikes a better balance, accurately restoring both the details and colors.
\begin{figure*}[t]
	\centering
	\includegraphics[width=2.1\columnwidth]{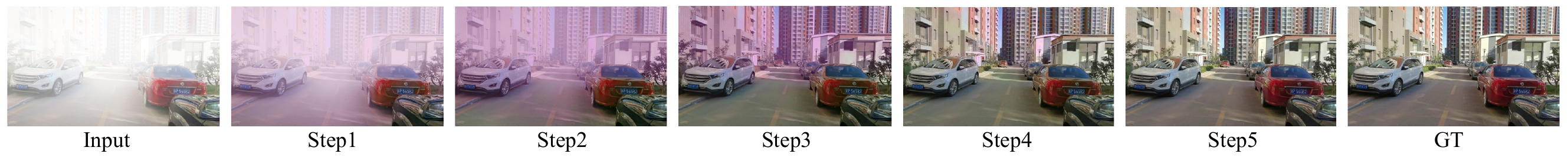} 
	\vspace{-4mm}
	\caption{Visualization of the dehazing process with the flow.
	}
	\vspace{-5mm}
	\label{figure-RK4}
\end{figure*}
\begin{figure}[t]
	\centering
	\includegraphics[width=\columnwidth]{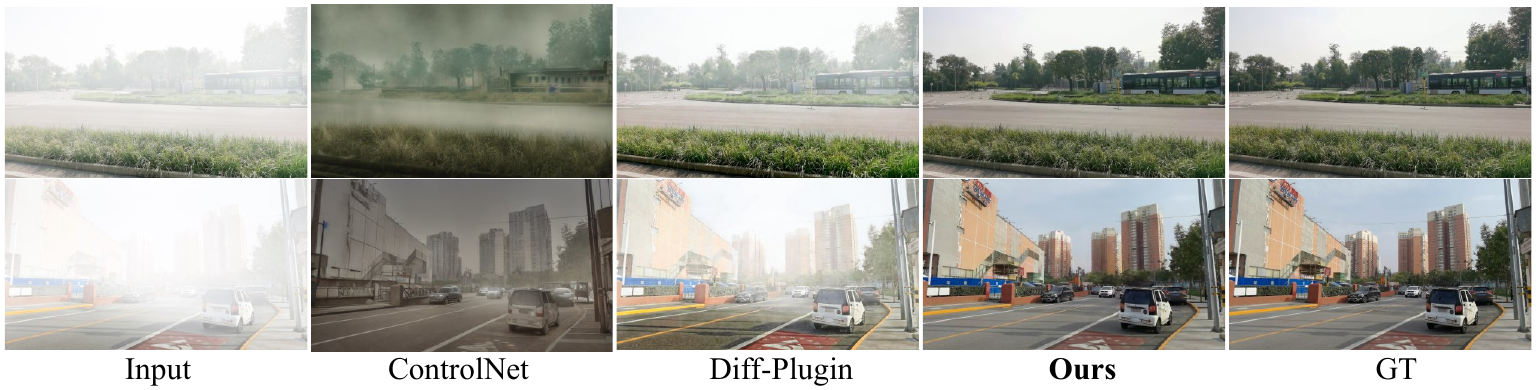}
	\vspace{-6mm}
	\caption{Visual comparison with diffusion-based methods.}
	\label{fig-diff}
	\vspace{-3mm}
\end{figure}
\subsection{Ablation study}
We perform ablation studies to demonstrate the effectiveness of the key components of our 4KDehazeFlow approach. 
For these experiments, we use the same architecture and hyperparameters and only vary one component for each ablation.

\noindent \textbf{Effectiveness of Haze-LUT.}
To verify the effectiveness of Haze-LUT, we conduct two ablation experiments by removing Haze-LUT and replacing it with a fixed LUT generated by contrast and saturation enhancement on normalized RGB values. 
As shown in Table \ref{table-ab}, removing Haze-LUT significantly degrades performance across all metrics, while the fixed LUT yields even lower performance, highlighting the importance of adaptive color correction. 
Figure~\ref{figure-ab}(a)-(d) show that the 
absence of Haze-LUT tends to cause color deviations, leading to over-enhancement in the blue channel.

\noindent \textbf{Effectiveness of $\boldsymbol{\lambda}$.}
We conduct ablation experiments on the 4KID dataset to investigate the influence of the parameter $\lambda$ in haze-aware vector field.
As shown in Table~\ref{table-ab}, setting $\lambda=0.1$ yields better performance than $\lambda=1$, but still produces suboptimal results compared to $\lambda=0.5$. 
%

\noindent \textbf{Effectiveness of RK4 ODE Solver.}
To evaluate the impact of the ODE solver, we conduct ablation experiments on the SOTS dataset using Euler~\cite{Euler}, Midpoint~\cite{midpoint}, and RK4 solvers. 
%
%
As shown in Table~\ref{table-ab}, RK4 achieves the best structural and perceptual quality. 
Figure~\ref{figure-ab} shows that Euler (e) and Midpoint (f) solvers lead to excessive blue brightness and red channel deviations from the GT (b), while RK4 (c) achieves better color accuracy, closer to the GT.

\section{Discussion}
\noindent \textbf{Real-World UHD image dehazing.}
We evaluate the proposed method on real-world hazy images. 
Figure~\ref{figure-real} shows the qualitative comparison of results on two challenging real-world images. 
As shown, Dehamer distorts the sky, causing unnatural transitions, while C2PNet and 4KDehazing result in overly dark and red-tinted images, leading to a significant degradation in color fidelity.  
UHDformer and UHDDIP introduce visible artifacts, degrading image quality. 
%
See the Supplementary Material for more results.

\begin{table}[t]
	\centering
	\small
	\caption{Quantitative comparison with diffusion-based methods. KID values ×100 for readability.}
	\vspace{-3mm}
	\resizebox{0.47\textwidth}{!}{
		\begin{tabular}{l|c c c c}
			\toprule
			\textbf{Methods} & \textbf{FID$\boldsymbol{\downarrow}$} & \textbf{KID$\boldsymbol{\downarrow}$} & \textbf{MUSIQ$\boldsymbol{\uparrow}$}& \textbf{TIME$\boldsymbol{\downarrow}$}\\
			\hline
			ControlNet~\cite{ControlNet}& 52.01 &13.54 &32.17 &8.94s \\
			Diff-Plugin~\cite{Diff-Plugin}& 47.58 &11.15 &36.55 &450.89s \\
			\textbf{4KDehazeFlow (Ours)}&\textbf{45.58}  &\textbf{8.33}  &\textbf{38.39} &\textbf{0.1540s}\\
			\bottomrule
		\end{tabular}
	}
	\label{table-diff}
	\vspace{-5mm}
\end{table}
\noindent \textbf{Flow path visualization.}
As shown in Figure~\ref{figure-RK4}, we demonstrate the step-by-step visualization of the progressive enhancement of a pair of 4K blurred images under our algorithm. We find that in the initial few steps, our method mainly focuses on image dehazing, while in the subsequent steps, it mainly restores the color and texture of the images.

\noindent \textbf{Comparison with diffusion-based methods.}
We compare our method with the diffusion-based Diff-Plugin~\cite{Diff-Plugin} and ControlNet~\cite{ControlNet}. 
The evaluation is conducted using FID, KID, and the no-reference MUSIQ metric. 
As shown in Table~\ref{table-diff}, our method outperforms both Diff-Plugin and ControlNet across all metrics on the 4KID dataset. 
Notably, 4KDehazeFlow processes a single 4K image in just 0.154 seconds on an A100 GPU, which is significantly faster than Diff-Plugin's 450.89 seconds and also much more efficient than ControlNet. 
Figure~\ref{fig-diff} shows that Diff-Plugin leaves haze residues, while ControlNet suffers from color distortions. 
Our method produces visually cleaner results with better color fidelity.
\section{Conclusion}
In this paper, we propose a UHD image dehazing method based on Flow Matching, which incorporates a dehazing prior assumption and a color restoration scheme. 
From the perspective of interpretability, the model is designed to first remove the haze in the image, and then refine the visual quality by restoring accurate and vivid color information. 
Our method achieves a good trade-off between accuracy and speed in reconstructing a clear UHD image.

\bibliography{aaai2026}

\end{document}